\documentclass[conference]{IEEEtran}
\IEEEoverridecommandlockouts
\usepackage{cite}
\usepackage{amsmath,amssymb,amsfonts}
\usepackage{algorithmic}
\usepackage{graphicx}
\usepackage{textcomp}
\usepackage{xcolor}
\usepackage{array} 
\usepackage{booktabs} 
\usepackage{caption} 
\usepackage{fancyhdr}

\fancypagestyle{FirstPage}{

\lfoot{979-8-3503-7903-7/24/\$31.00 ~\copyright2024 IEEE}

}
\def\BibTeX{{\rm B\kern-.05em{\sc i\kern-.025em b}\kern-.08em
    T\kern-.1667em\lower.7ex\hbox{E}\kern-.125emX}}
\begin{document}

\title{Developing Normative Gait Cycle Parameters for Clinical Analysis Using Human Pose Estimation 
}
\author{
Rahm Ranjan$^{1,2}$,
David Ahmedt-Aristizabal$^{1}$, 
Mohammad Ali Armin$^{1}$,
Juno Kim$^{2}$\\
$^{1}$ CSIRO Data61, Australia. \\
$^{2}$ University of New South Wales, Australia.\\
{\tt\small \{rahm.ranjan\}@unsw.edu.au}\\
}

\maketitle



\begin{abstract}
Gait analysis using computer vision is an emerging field in AI, offering clinicians an objective, multi-feature approach to analyse complex movements. 
Despite its promise, current applications using RGB video data alone are limited in measuring clinically relevant spatial and temporal kinematics and establishing normative parameters essential for identifying movement abnormalities within a gait cycle. 
This paper presents a data-driven method using RGB video data and 2D human pose estimation for developing normative kinematic gait parameters. By analysing joint angles, an established kinematic measure in biomechanics and clinical practice, we aim to enhance gait analysis capabilities and improve explainability. 
Our cycle-wise kinematic analysis enables clinicians to simultaneously measure and compare multiple joint angles, assessing individuals against a normative population using just monocular RGB video. This approach expands clinical capacity, supports objective decision-making, and automates the identification of specific spatial and temporal deviations and abnormalities within the gait cycle. 
\end{abstract}

\begin{IEEEkeywords}
Gait, Clinical Gait Analysis, Human Pose Estimation, Gait Parameters.
\end{IEEEkeywords}

\section{Introduction}
Human gait analysis is critical in healthcare, robotics, biometrics, and surveillance. With advancements in computer vision and machine learning, video-based human pose estimation and action recognition have emerged as powerful tools for gait analysis, offering usability and utility for healthcare and clinical applications~\cite{sethi2022}. 
However, existing computer vision techniques for clinical gait analysis often fall short in meeting the clinical requirements of identifying spatial and temporal abnormalities using RGB video data. 
This is primarily due to limited annotated RGB video datasets and the absence of well-defined normative kinematic parameters. Further, many feature-engineering approaches are highly specific to gait and lack generalisability to other actions. Therefore, there is a need for a more versatile and clinically relevant method that assists clinicians in detecting and visualising movement abnormalities across diverse actions and conditions.

Current computer vision research involving gait is focused on action classification~\cite{kong2022}, person identification / gait recognition~\cite{Sepas-Moghaddam2023}, and gait pathology classification~\cite{goyal2020detection}. While these tasks are important, they often fall short in providing clinically important information, such as the spatial and temporal localisation of movement abnormalities\cite{roberts2017biomechanical}. 

In clinical practice, identifying normal kinematics is routine for detecting spatial and temporal issues in a person's gait~\cite{baker2006gait}. Typically, clinical experts are trained to visually identify body parts and joints that demonstrate atypical movement patterns. However, translating this routine practice remains challenging in computer vision for several reasons. Two primary reasons include the limited availability of accurately annotated video data and the lack of well-defined normative gait parameters. 

As a result, researchers have approached this issue using a mix of self-derived features and clinically established gait parameters. Novel features include pixel joint locations, kinematic measurements, and mixed spatial and temporal features, which are typically feature-crafted metrics selected for gait analysis tasks. However, these feature-crafted approaches often fail to generalise to other sub-classes within the action and are limited when applied to other actions. For example, Goyal et al.~\cite{goyal2020detection} create non-conventional crafted features such as ``limb-straightness'' and ``hand-leg coordination'' to classify neurological gait styles. However, this approach can only be utilised for gait classification and is limited with respect to non-gait actions such as throwing a ball. Even approaches that use conventional gait parametric features \cite{Jun2023Hybrid}, such as step length and gait velocity, have demonstrated utility only for gait actions. The reliance on gait-specific parameters limits the scope of action analysis so it may detect non-gait actions with reference only to gait. As a result, using this method to determine abnormalities in non-gait actions is likely to be inaccurate. In the case of normal or efficient ball-throwing action against abnormal or inefficient ball-throwing action, clinicians and sporting coaches will use upper body joint positions and angles instead of step length.

To overcome these limitations, we provide the following contributions:
1) A novel, clinically relevant method for detecting and visualising potential kinematic abnormalities during gait. 
2) A baseline framework for identifying kinematic abnormalities in gait analysis.
3) Introduction of clinical normative values for gait using universal kinematic measures inferred from RGB video data.
4) A cycle-wise clinical annotation for tracking multiple joint movements simultaneously.

\thispagestyle{FirstPage}

\section{Background on Clinical Gait Analysis}

Human gait analysis is an important diagnostic tool used in clinical settings to assess musculoskeletal and neurological conditions affecting human movement. Clinical gait analysis (CGA) involves the systematic measurement, evaluation, and interpretation of gait parameters to gain insights into a person's walking patterns and identify abnormalities or impairments~\cite{baker2006gait}. This analysis provides quantitative and qualitative information about the kinematics, kinetics, and muscle activity during walking, enabling healthcare professionals to develop targeted treatment plans and monitor progress \cite{whittle2014gait}. 

The primary objectives of CGA are to identify deviations from typical gait patterns and understand the mechanisms causing these deviations. Many internal and external factors contribute to abnormalities in gait~\cite{chang2010role}. External factors such as the environment, terrain, clothing, and carried items significantly influence an individual's gait pattern~\cite{baker2006gait}. However, most clinical gait analysis focuses on internal factors such as range of motion, muscle strength and flexibility~\cite{whittle2014gait}, as well as neurological~\cite{jahn2010gait} and psychological factors~\cite{lucyyardley2004psychosocial}. Clinicians will typically observe the gait action within a controlled environment such as a clinic or home as an initial screening assessment~\cite{baker2006gait}.  

Clinical references for typical human movements are used to classify gait normality in the clinical setting~\cite{roberts2017biomechanical}. There is no universal metric for this task; instead, many gait parameters with reference norms exist for various purposes and conditions~\cite{whittle2014gait}. 
Clinicians observe abnormalities in joint angles, stride length, step width, and other parameters to aid clinical decision-making. Often, a secondary parameter such as gait speed is used to simplify the multifaceted nature of complex movement, making it simpler for clinicians to understand. However, this simplification can result in the loss of vital information useful for identifying the cause of movement abnormalities. 

Deviations from typical ranges of movement can indicate underlying conditions such as muscle weakness~\cite{sicard2002gait}, joint stiffness~\cite{totah2019impact}, balance issues~\cite{baker2006gait}, or neurological disorders~\cite{jahn2010gait}. The analysis can also identify compensatory movements that patients may adapt to accommodate their impairments~\cite{whittle2014gait}, providing valuable insights into functional limitations and potential injury risks. The location and timing of abnormal movement provide more information regarding the underlying nature of the issue and potential solutions~\cite{baker2006gait}. This information guides the development of personalised treatment plans, which may involve rehabilitation exercises, orthotic devices, surgical interventions, or other interventions tailored to address the specific impairments identified.

Typically, objective and accurate clinical gait analysis requires specialised equipment to capture motion and force\cite{wren2011efficacy}. Sensors, optical devices, force plates, electromyography (EMG), and other instruments are used to capture and measure various aspects of gait. These technologies enable the collection of accurate and objective data, allowing for quantitative assessment and comparison of gait parameters across different individuals and time points\cite{di2020gait}.

Access to specialised equipment for clinical gait analysis is often challenging for many healthcare professionals due to various factors such as cost and geographical constraints~\cite{krebs1985reliability}. The high cost of sophisticated gait analysis systems, including marker-based motion capture systems, force plates, and EMG devices, can pose a significant barrier for smaller clinics or healthcare facilities with limited budgets. Furthermore, these specialised instruments require technical expertise for setup, calibration, and data interpretation, which may not be readily available to all clinicians~\cite{krebs1985reliability}.
As a result, many clinicians resort to alternative methods and tools, such as visual observation, functional assessments, and video-recorded analysis to assess gait and obtain valuable information about a patient's walking patterns~\cite{roberts2017biomechanical}. However, these alternative methods typically have reduced accuracy compared to sensor or marker-based motion capture systems. Video capture assists with clinical human movement analysis and is considered the best clinical practice~\cite{wiles2003use}, given the availability of integrated cameras and storage in mobile computing devices. Video-based analysis has been validated as a less accurate but clinically useful alternative to more expensive marker-based optical motion capture systems, with the added benefit of usability outside of the clinical environment~\cite{Nakano2020Evaluation}.

\section{Related Works}

\noindent\textbf{Computer Vision Methods for Clinical Gait Analysis.}
%
Computer vision methods are gaining prominence in healthcare, offering cost-effective, user-friendly, objective, and automated analysis for various health-related tasks, including clinical gait analysis (CGA)~\cite{Milstein2020Computer}. These techniques extract gait parameters to classify pathologies such as Parkinson's disease, Huntington's disease, cerebral palsy, and stroke, utilising data from 3D motion capture systems, RGB-D sensors, or common RGB cameras. CGA is distinctively different from other computer vision gait related tasks in that gait is classified according to gait pathology type, gait pathology severity, and measures of kinematics. 
Recent studies have employed diverse deep learning approaches. Recurrent Neural Networks (RNNs) such as Long Short-Term Memory (LSTM) models have been used to detect gait events in children with cerebral palsy~\cite{Kidzinski2019Automatic, Kidzinski2020Deep}, identify gait abnormalities using skeletal data \cite{Khokhlova2019Normal}, and accurately identify gait events across various walking patterns~\cite{Kim2022deeplearning}. 
Convolutional Neural Networks (CNNs) have demonstrated kinematic analysis using monocular RGB video~\cite{Kidzinski2020Deep}. Graph Convolutional Networks (GCNs) have been proposed for classifying vestibular pathologies~\cite{Tian2022Skeletonbased} and pathological gaits~\cite{Kim2022PathologicalGait}, with some approaches providing explainable outputs through techniques such as class activation maps.
These studies have shown the potential of computer vision in automating various aspects of gait analysis, including the detection of gait events and cycles, estimation of spatiotemporal gait parameters and joint angles~\cite{Stenum2021Twodimensional}, and classification of various pathological gaits. Hybrid approaches have fused features into unified feature vectors from GCNs, RNNs, and basic gait parameters to classify neurological pathologies~\cite{Jun2023Hybrid}, while transformer models have also been applied to gait pattern classification~\cite{cheriet2023multi}.
Challenges remain, including limited publicly available datasets of pathological gaits and the impact of camera positioning on spatial measurements. Research has also indicated that the effectiveness of different features (joint angles vs. gait parameters) varies depending on the specific pathology being classified, with some conditions relying more on joint angle sequences and others on basic gait parameters for accurate classification~\cite{Jun2023Hybrid}. Despite these challenges, computer vision methods continue to advance, promising more efficient and accurate gait analysis techniques for clinical applications.

\noindent\textbf{Computer Vision Involving Gait Actions.}
%
Computer vision has been well-established for the tasks of gait recognition, person identification and human action recognition. These tasks are primarily grounded in surveillance and biometrics~\cite{Harris2022Survey}. Gait recognition involves using gait as a biometric feature to identify a person based on their body shape, posture, and walking pattern. Similar to other biometrics such as fingerprints, a gait pattern is unique to an individual and has the advantage of being identifiable at a distance~\cite{FilipiGoncalvesDosSantos2023Gait}. Gait recognition is primarily concerned with person identification; however, there are sub-tasks such as person re-identification~\cite{Harris2022Survey}, emotion detection\cite{Sheng2021multi,Bhattacharya2020STEP}, age estimation, and gender recognition. Datasets provided for these tasks may contain gait actions however often have been unsuitable or limited for use with CGA~\cite{Ranjan2024GAVD}. Furthermore, these datasets often lack clinical annotations required for clinically validated gait analysis. 

Given the limited approaches using computer vision for clinical gait analysis with RGB Video data, gait recognition methods offer many techniques that are transferable for aspects of gait analysis. However, it should be noted that the gait recognition task is mainly concerned with classification and lacks progress towards explainability. To address this, we propose a method for articulating the baseline parameters of the normal human gait cycle so that potential sources of deviation relative to abnormal gait can be spatially and temporally referenced to specific phases of the gait cycle.

\section{Method}

\subsection{Video Datasets}
RGB videos consisting of gait action sequences are acquired from three different data sources (See Table~\ref{tab:gait_vid_source}). 
The first is the popular GPJATK~\cite{Kwolek2019Calibrated}, the second includes data from the Gait Abnormality in Video Dataset (GAVD)~\cite{Ranjan2024GAVD}, and 
the third is from the Clinical Abnormality Simulated Dataset (CASD)~\cite{Ranjan2024GAVD}.


\subsubsection{GPJATK gait dataset}
The GPJATK dataset~\cite{Kwolek2019Calibrated} is a video dataset captured in a controlled indoor environment. Subjects have no observable gait abnormalities as determined by an expert clinician. We utilise only a selection of normal gait sequences taken from the left side view perspective. This includes 42 normal gait sequence videos with 42 different subjects demonstrating 99 complete normal gait cycles.

\subsubsection{Gait Abnormality in Video Dataset (GAVD)}
The GAVD dataset~\cite{Ranjan2024GAVD} consist of gait videos from in-the-wild. These videos are screened by clinicians and sourced from online video-sharing platforms. For this study, both normal and abnormal gaits are collected and annotated by clinicians. However, we utilise only a selection of normal gait sequences taken from the left side view perspective for the development of normal gait parameters, in accordance with videos from other data sources. This includes 11 normal gait sequence videos with 11 different subjects demonstrating 71 complete normal gait cycles.

\subsubsection{Clinical Abnormality Simulated Dataset (CASD)}
Controlled gait videos~\cite{Ranjan2024GAVD}, simulating clinical gait analysis, are captured with 20 subjects walking in controlled indoor conditions supervised by healthcare experts. Subjects walk multiple laps in a straight line with left side views recorded. Subjects perform four different gait styles instructed by clinicians consisting of both normal and abnormal gait styles. For consistency, we utilise only a selection of normal gait sequences taken from the left side view perspective to develop the gait parameters. From 20 videos with 20 subjects, there are 137 complete normal gait cycles performed.

\begin{table}[t]
\caption{Gait Video Sample Counts for Gait Parameters }
\vspace{-6pt}
\begin{center}
\begin{tabular}{|c|c|c|c|}
\hline
\textbf{Dataset} & \textbf{\textit{Subjects}}& \textbf{\textit{Gait Videos}}& \textbf{\textit{Cycles}} \\
\hline
GPJATK~\cite{Kwolek2019Calibrated}& 11 & 11& 71 \\
\hline
GAVD~\cite{Ranjan2024GAVD}& 42 & 42& 99 \\
\hline
CASD~\cite{Ranjan2024GAVD}& 20& 20& 181 \\
\hline
\textbf{Totals}& \textbf{73}& \textbf{73}& \textbf{351} \\
\hline
\end{tabular}
\label{tab:gait_vid_source}
\vspace{-10pt}
\end{center}
\end{table}

\subsection{Annotation}
An expert clinician screened and annotated relevant videos, ensuring clinically accurate identification of gait events and typical gait patterns. Previous studies~\cite{Toro2003review,Lee2020Perceiving} have established the reliability and consistency of expert clinicians, such as physiotherapists, to accurately identify typical and atypical gait events. The LabelBox tool~\cite{labelbox} is used to annotate the start and end of each gait cycle, both defined in this study as the moment of the left foot's initial heel strike. Each gait cycle is further labelled with clinically relevant gait of typical or atypical.  

\subsection{Feature Extraction}
BlazePose~\cite{Bazarevsky2020BlazePose} human pose estimation model (HPE) is used to extract key point features. It is selected due to its lightweight deep learning architecture that balances accuracy with computational efficiency. BlazePose is validated as viable tool for certain gait analysis applications such as CGA in controlled environments~\cite{Chakraborty2021Validation}. BlazePose has also been clinically validated for exercise related action measurement~\cite{Capecci2021Real}. 
A local computer with a GPU (NVIDIA GeForce RTX 3080) is used to run the HPE model to ensure that videos containing identifiable participant information are not shared with any third-party services. Each frame from the gait video sequences is used as input for the HPE model.
Fourteen (14) keypoints are extracted from each frame as pixel coordinates, including  bilateral keypoints of the shoulders, elbows, wrists, hips, knees, ankles, heels, and hallux. In this study, camera calibration and distortion correction were not available; this limitation is addressed in Section~\ref{subsect:limits_future_work}.

\begin{table}[t]
    \centering
    \caption{Keypoints Used for Joint Angles}
    \label{Tab:joint_angles}
    \resizebox{0.48\textwidth}{!}{%
    \begin{tabular}{lcccc}
        \toprule
        \textbf{Joint Angle} & \textbf{Proximal Keypoint} & \textbf{Axis Keypoint} & \textbf{Distal Keypoint} \\
        \midrule
        (L) Shoulder & (L) Hip & (L) Shoulder & (L) Elbow \\
        (R) Shoulder &(R) Hip &(R) Shoulder &(R) Elbow \\
        \midrule
        (L) Elbow&(L) Shoulder &(L) Elbow &(L) Wrist \\
        (R) Elbow &(R) Shoulder &(R) Elbow &(R) Wrist \\
        \midrule
       (L) Hip &(L) Shoulder &(L) Hip &(L) Knee \\
       (R) Hip & (R) Shoulder & (R) Hip & (R) Knee \\
        \midrule
        (L) Knee&(L) Hip &(L) Knee &(L) Ankle \\
       (R) Knee  &(R) Hip &(R) Knee &(R) Ankle \\
        \midrule(L)
        Ankle &(L) Knee &(L) Ankle &(L) Hallux \\
        (R) Ankle&(R) Knee &(R) Ankle &(R) Hallux \\
        \bottomrule
        \multicolumn{4}{l}{(L) = Left; (R) = Right}
    \end{tabular}}
\vspace{-6pt}
\end{table}

\subsection{Joint Angle Calculation} 
Joint angles are selected as a universal kinematic feature providing great clinical utility as an objective measure for both gait and non-gait human actions. Joint angles are clinically recognised as outcome measures that normalise variations in subject size, allowing for comparisons between different subjects or across multiple observations of the same subject. 
We have selected ten joints commonly used in clinical gait analysis~\cite{whittle2014gait} (See Table~\ref{Tab:joint_angles}). 
Using a single lateral camera perspective, joint angles are computed in 2D, mimicking the clinical process of manually establishing joint angles from RGB video. 
To calculate the joint angle formed by three keypoints---Proximal Keypoint \(A\), Axis Keypoint \(B\), and Distal Keypoint \(C\)---we employ a vector-based method following Pagnon et al.~\cite{Pagnon2023}. 
Utilising the \texttt{arctan2} function from the NumPy library, the joint angle is computed as the angle between the vectors \( \overrightarrow{BA} \) and \( \overrightarrow{BC} \) relative to the horizontal axis. Specifically, the angle between vectors \( \overrightarrow{BC} \) and \( \overrightarrow{BA} \) is calculated as:
\begin{equation}
\text{radians} = \text{atan2}(c_y - b_y, c_x - b_x) - \text{atan2}(a_y - b_y, a_x - b_x)
\end{equation}
where \( (a_x, a_y) \), \( (b_x, b_y) \), and \( (c_x, c_y) \) represent the coordinates of points \( A \), \( B \), and \( C \), respectively. The result is then converted from radians to degrees, and the absolute value is taken to ensure the angle is positive:
\begin{equation}
\text{angle} = \left| \frac{\text{radians} \times 180.0}{\pi} \right|
\end{equation}
To ensure the angle falls within the conventional range of 0 to 180 degrees, any computed angle greater than 180 degrees is adjusted by subtracting it from 360 degrees. This approach guarantees that the calculated angle accurately reflects the geometric relationship between the three points.

\subsection{Creating Normative Parameters} 
\begin{figure}[!t]
    \centering
    \includegraphics[width=0.49\textwidth]{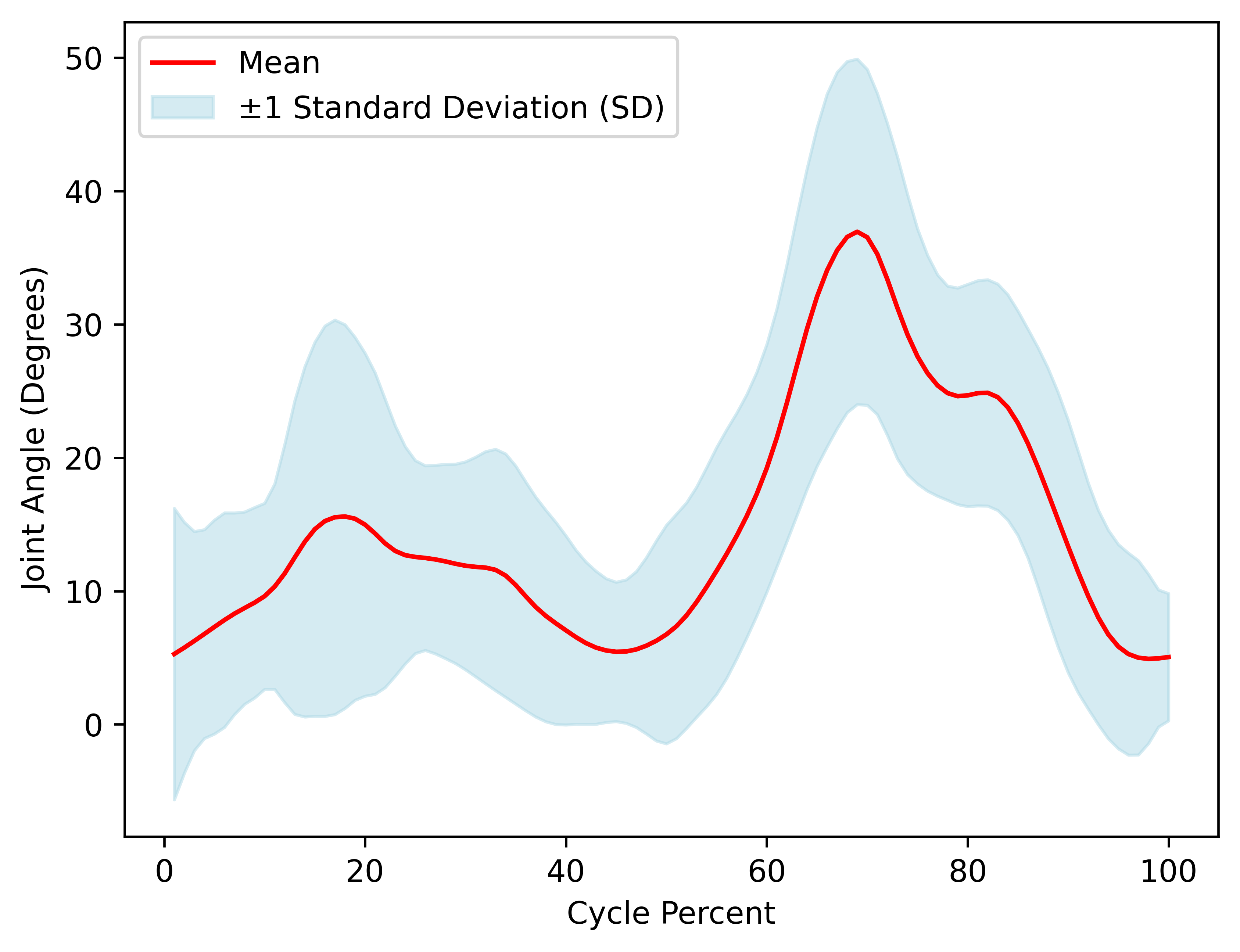} 
    \caption{Example of typical kinematic parameters of left knee during gait cycle extrapolated from video of 351 normalised gait cycles.}
    \label{fig:leftknee}
\end{figure}

A clinically representative normative parameter for each joint is derived from a large RGB video dataset of typical gait cycles (See Example Fig.~\ref{fig:leftknee}). 
Joint angle kinematic data is calculated for each joint and temporally normalised, with initial heel contact of the left leg marking the start and end of the gait cycle. The start of the cycle is designated as 0\% and the end of the cycle is 100\% of the gait cycle. We determine each frame's approximate position within this cycle. To handle missing data, cubic spline interpolation~\cite{DeBoor1978Practical} is applied using Scikit-learn library for Python~\cite{scikit-learn}. This ensures smooth joint angle estimates across the entire gait cycle. Consistent with current practise in gait analysis, such as the Conventional Gait Model~\cite{Baker2017Conventional}, statistical measures---mean and standard deviation---are calculated from the interpolated data, providing normative values and tolerances to determine typical joint angle measures.

\section{Experiments}

\subsection{Experimental Setup}
To demonstrate cycle-wise kinematic deviation and abnormality detection using our parameters, we test on four randomly selected unseen gait videos sourced from the various aforementioned datasets (See Table~\ref{tab:gait_vid_source}). Videos one and three consist of typical gait patterns, while videos two and four consist of atypical gait patterns. The subject in videos three and four is the same individual performing both typical and atypical gait patterns in the same recorded environment to demonstrate inter-subject kinematic differences detection (See Table~\ref{tab:test_vid}). 
We demonstrate single joint and multi-joint abnormality analysis, assess the severity of deviation from typical movement, and provide qualitative analysis of selected frames from visualisation of joint abnormality detection. 

\begin{table}[t]
\caption{Test Video Details}
\begin{center}
\vspace{-6pt}
\begin{tabular}{|c|c|c|c|}
\hline
\textbf{Video} & \textbf{\textit{Source Dataset}}& \textbf{\textit{Gait Pattern}}& \textbf{\textit{Subject}} \\
\hline
one& GPJATK~\cite{Kwolek2019Calibrated} & typical& 1 \\
\hline
two& GAVD~\cite{Ranjan2024GAVD} & atypical& 2 \\
\hline
three& CASD~\cite{Ranjan2024GAVD}& typical& 3 \\
\hline
four& CASD~\cite{Ranjan2024GAVD}& atypical& 3 \\

\hline
\end{tabular}
\label{tab:test_vid}
\vspace{-10pt}
\end{center}
\end{table}

\subsection{Results and Discussion}
In this paper, we demonstrate a data-driven approach to creating normative kinematic values for the clinically important action of gait (See Fig.~\ref{fig:leftknee}). These parameters are applied to detect possible deviations and abnormal joint angle kinematics for single joints and multiple joints and provide qualitative visualisations useful for clinical explanation of detected abnormalities.

\subsubsection{Single Joint Abnormality Detection}
\begin{figure}[!t]
    \centering
    \includegraphics[width=0.49\textwidth]{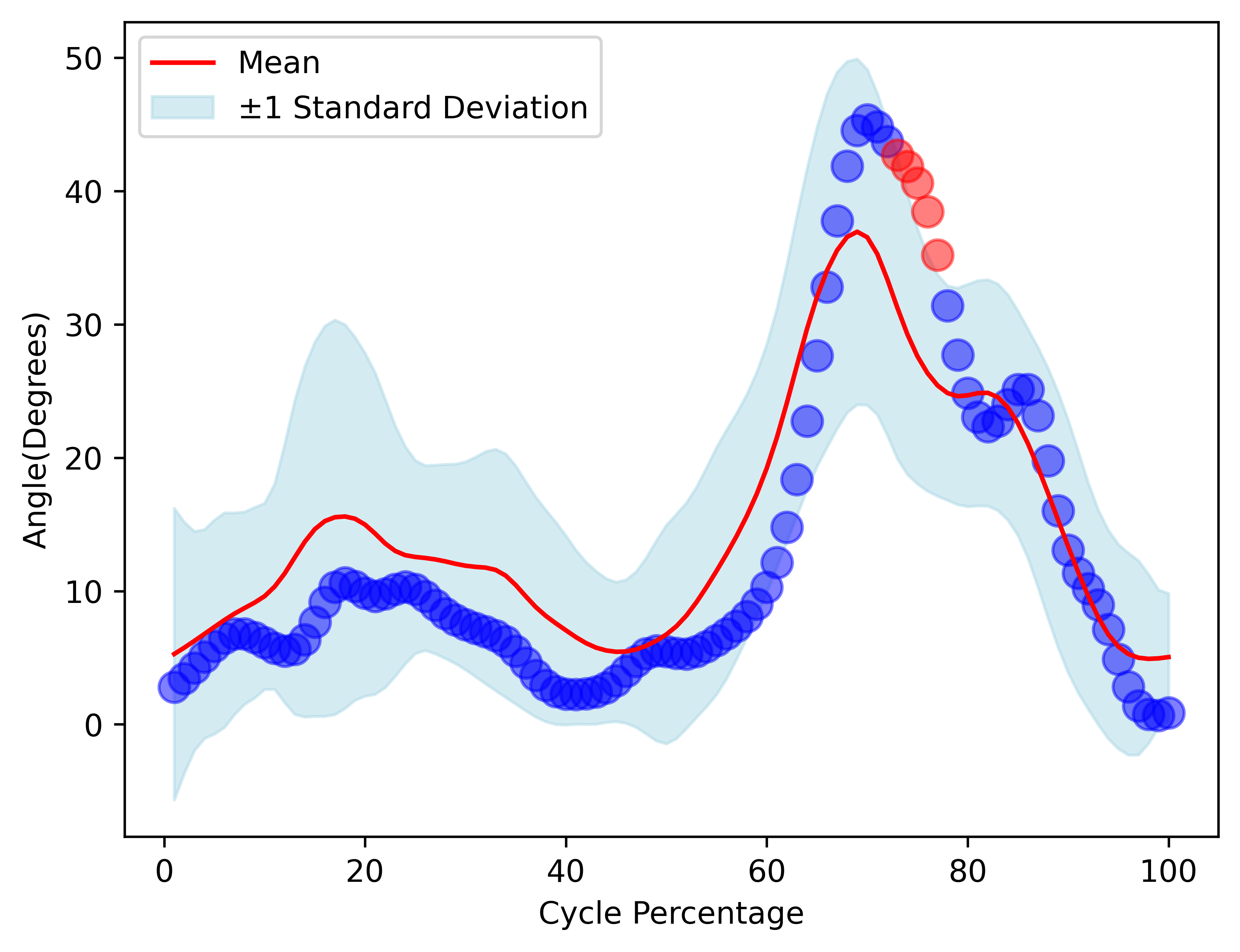} 
    \caption{Single joint detection of potential deviation of left knee angle during gait cycle, blue dots represent normal and red dots indicate abnormal.}
    \label{fig:Single Joint Abnormality Detection}
\vspace{-6pt}    
\end{figure}

Single Joint analysis reveals the detection of potentially abnormal joint angles throughout the gait cycle. Fig.~\ref{fig:Single Joint Abnormality Detection} shows the typical (blue) and atypical (red) detected points. Using the constructed gait parameters for the left knee joint, potential abnormal joint angles are identified as those falling outside of one standard deviation from the mean. The use of one standard deviation is valuable for clinicians in determining the specific joint deviations when compared to a representative population with typical movement. 

\begin{figure}[!t]
    \centering
    \includegraphics[width=0.49\textwidth]{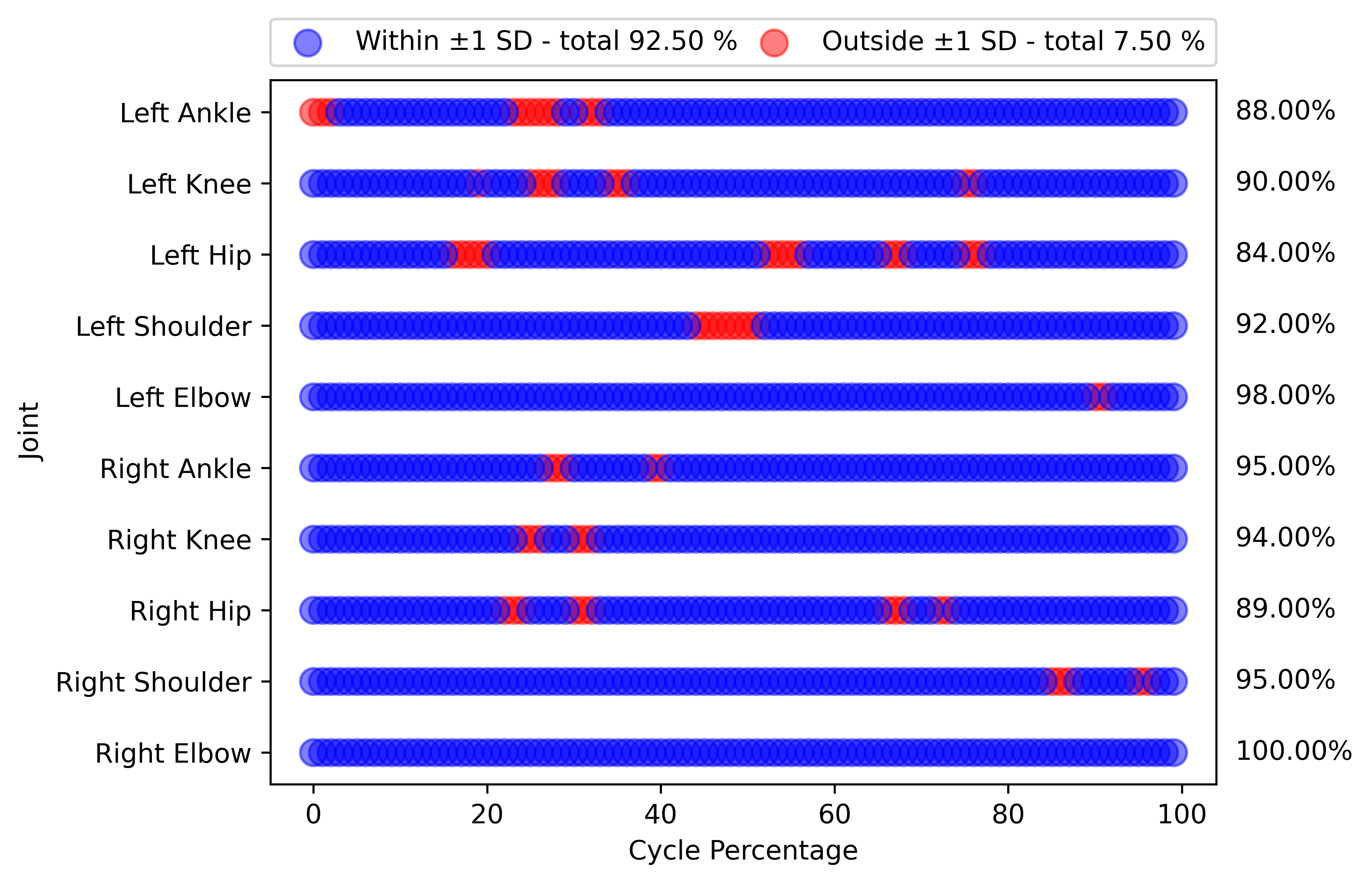} 
    \caption{Multi-joint abnormality detection for test video \textbf{one from GPJATK dataset}~\cite{Kwolek2019Calibrated} with \textbf{typical} gait pattern.}
    \label{fig:multi_vid_1}
\end{figure}

\subsubsection{Multi-Joint Abnormality Detection}
\begin{figure}[!t]
    \centering
    \includegraphics[width=0.49\textwidth]{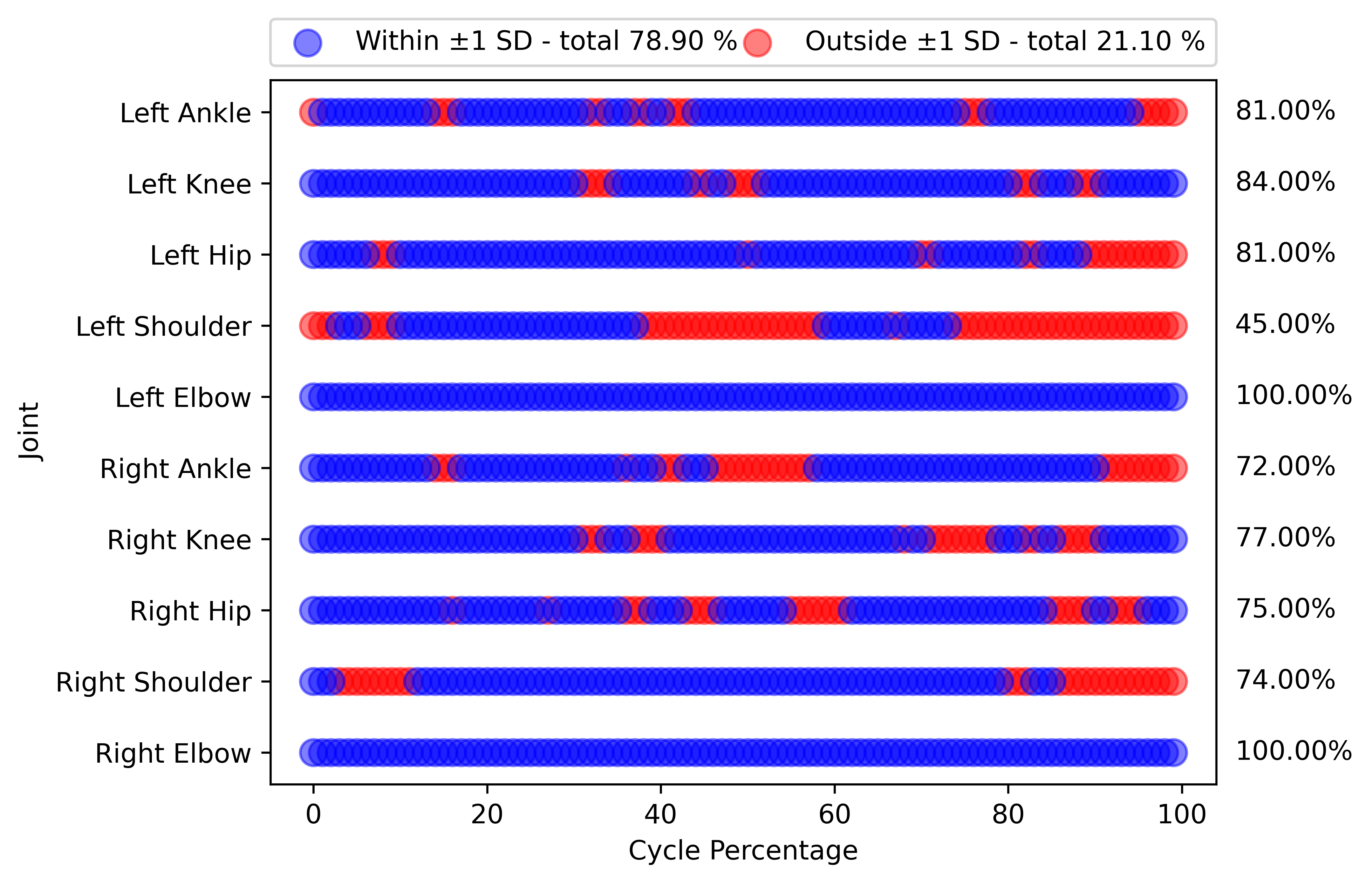} 
    \caption{Multi-joint abnormality detection for test video \textbf{two from GAVD}~\cite{Ranjan2024GAVD} with \textbf{atypical} gait pattern}
    \label{fig:multi_vid_2}
\end{figure}

\begin{figure}[!t]
    \centering
    \includegraphics[width=0.49\textwidth]{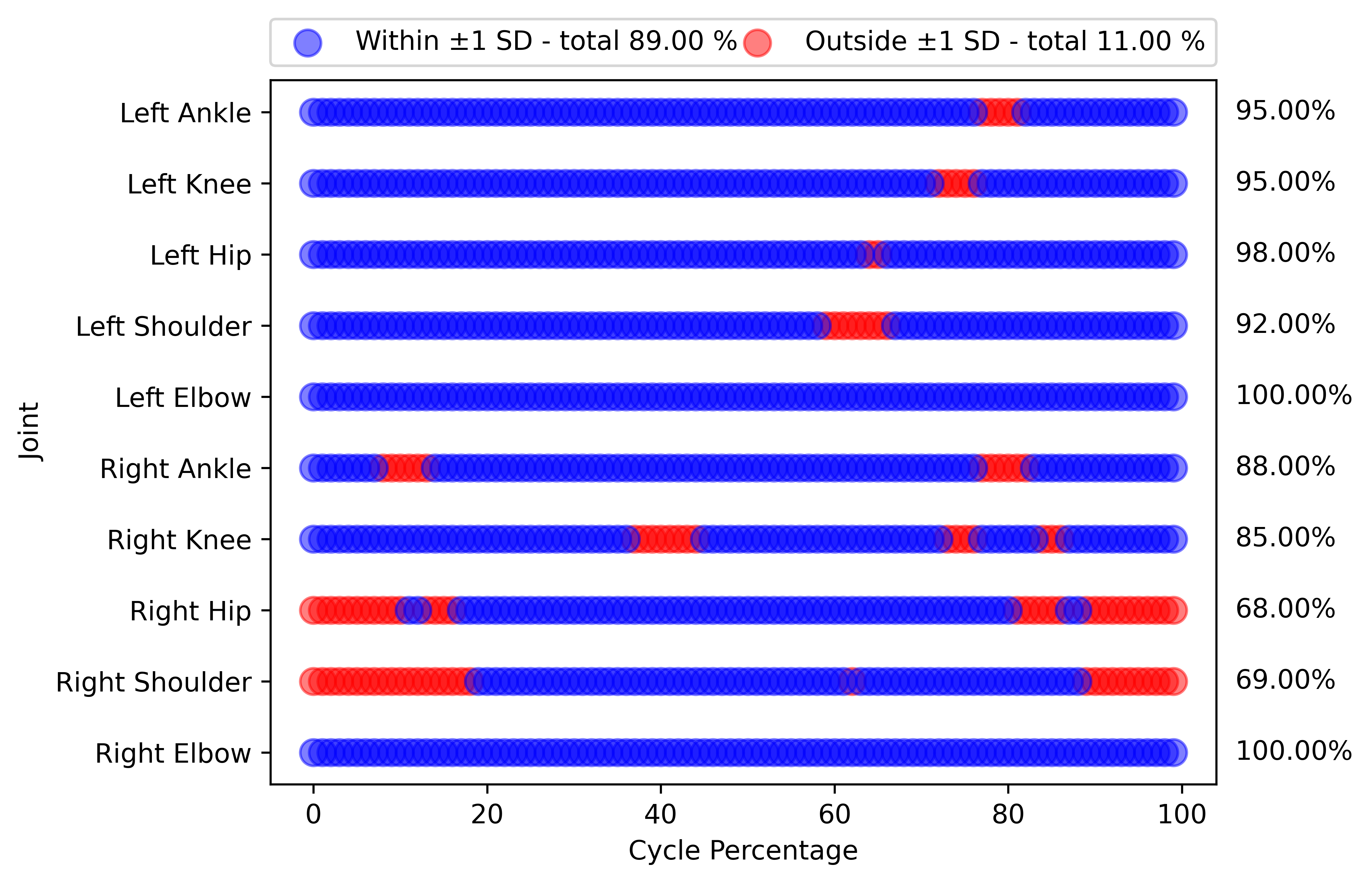} 
    \caption{Multi-joint abnormality detection for test video \textbf{three from CASD}~\cite{Ranjan2024GAVD} with \textbf{typical} gait pattern.}
    \label{fig:multi_vid_3}
\end{figure}

\begin{figure}[!t]
    \centering
    \includegraphics[width=0.49\textwidth]{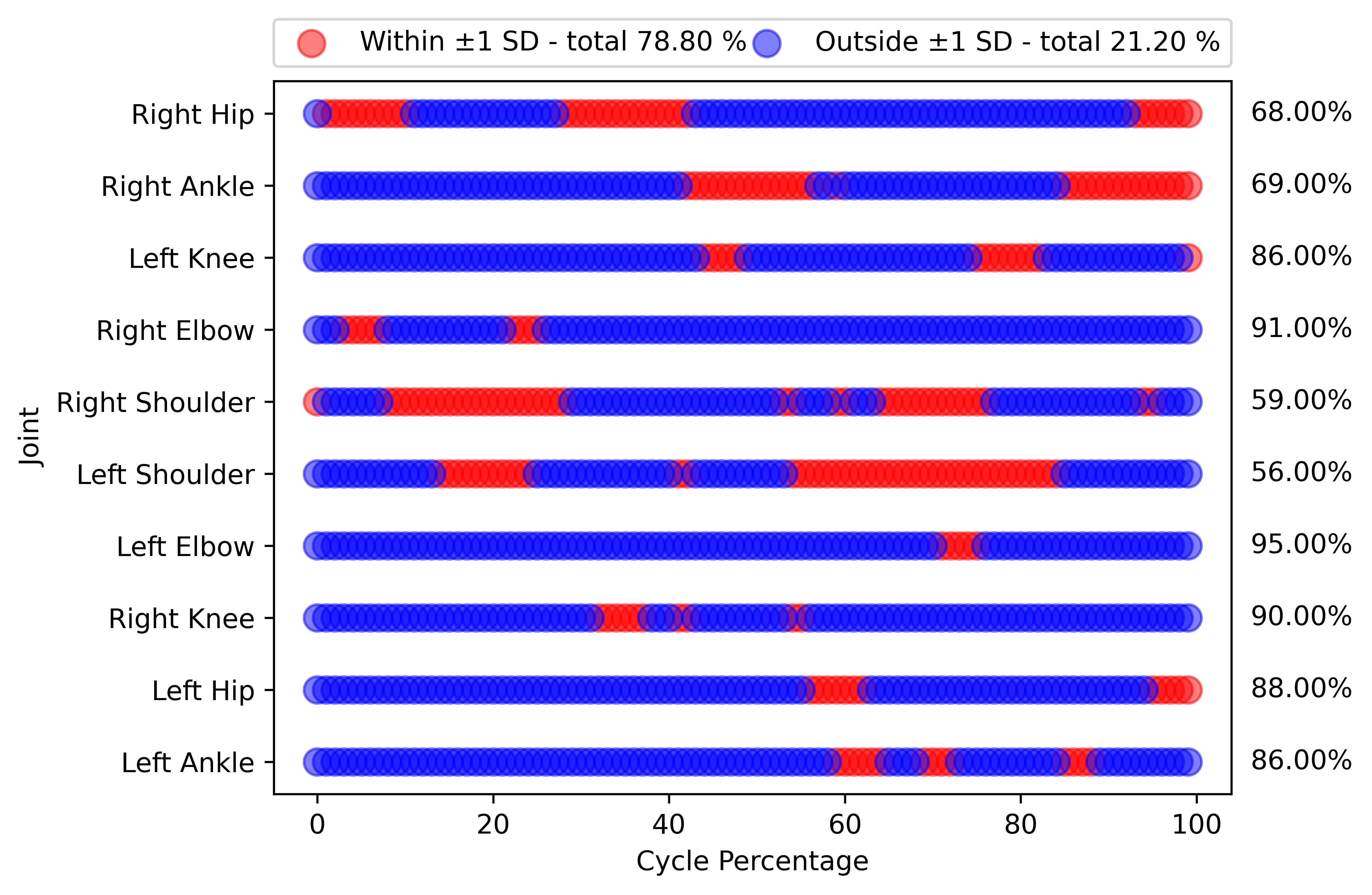} 
    \caption{Multi-joint abnormality detection for test video \textbf{four from CASD}~\cite{Ranjan2024GAVD}with \textbf{atypical} gait pattern.}
    \label{fig:multi_vid_4}
\end{figure}

We apply the detected points of typical (blue) and atypical (red) joint angles for multiple joints throughout the gait cycle to demonstrate the ability of this approach to analyse and highlight multiple outputs, a task that is challenging for clinicians to perform manually. 
The results of this analysis clearly demonstrate the increased detection of potential abnormalities.  Fig.~\ref{fig:multi_vid_1} and Fig.~\ref{fig:multi_vid_3} show that with typical gait patterns, minimal abnormality in joint angles is detected. In contrast, an observable increase in joint angles found outside of one standard deviation for typical gait kinematic parameters is seen with atypical gait patterns, as shown in Fig.\ref{fig:multi_vid_2} and Fig.~\ref{fig:multi_vid_4}


\begin{figure*}[!h]
    \centering
    \includegraphics[width=0.98\textwidth]{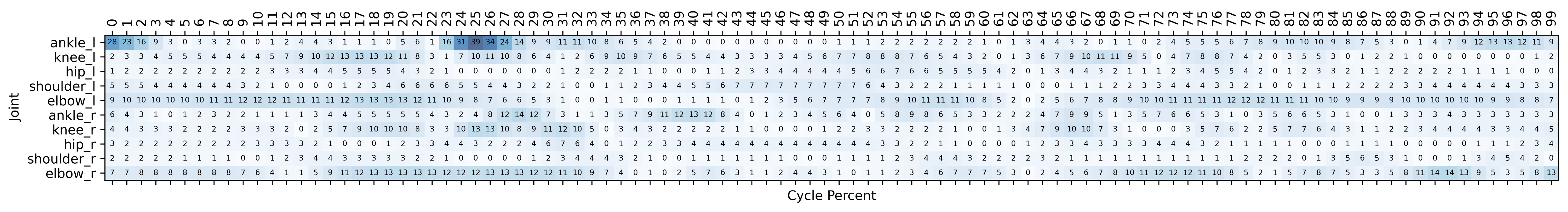} 
    \caption{Degree of deviation from typical for test video \textbf{one from GPJATK dataset}~\cite{Kwolek2019Calibrated} with \textbf{typical} gait pattern. Dark regions indicating greater degree of deviation from mean angle for cycle percent}
\vspace{-6pt}
    \label{fig:vid_1_matrix}
\end{figure*}

\begin{figure*}[!h]
    \centering
    \includegraphics[width=0.98\textwidth]{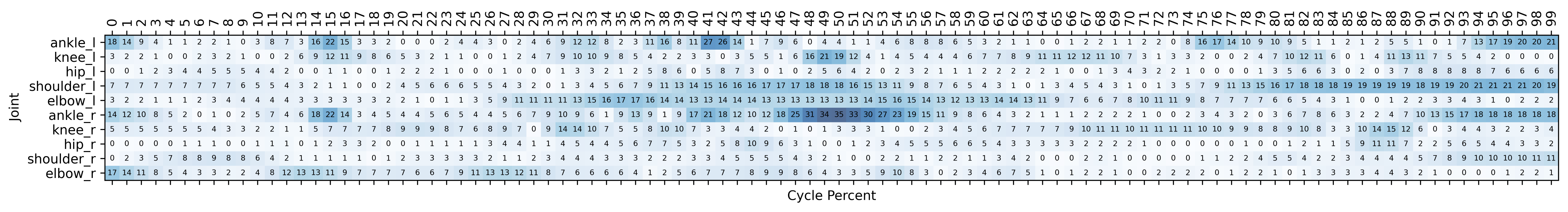} 
    \caption{Degree of deviation from typical for test video \textbf{two from from GAVD}~\cite{Ranjan2024GAVD} with \textbf{atypical} gait pattern. Dark regions indicating greater degree of deviation from mean angle for cycle percent}
\vspace{-6pt}
    \label{fig:vid_2_matrix}
\end{figure*}

\begin{figure*}[!h]
    \centering
    \includegraphics[width=0.98\textwidth]{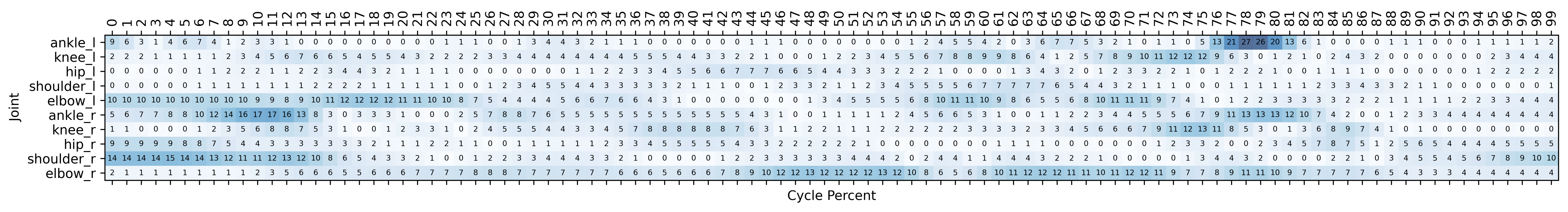} 
    \caption{Degree of deviation from typical for test video \textbf{three from CASD}~\cite{Ranjan2024GAVD} with \textbf{typical} gait pattern. Dark regions indicating greater degree of deviation from mean angle for cycle percent}
\vspace{-6pt}
    \label{fig:vid_3_matrix}
\end{figure*}

\begin{figure*}[!h]
    \centering
    \includegraphics[width=0.98\textwidth]{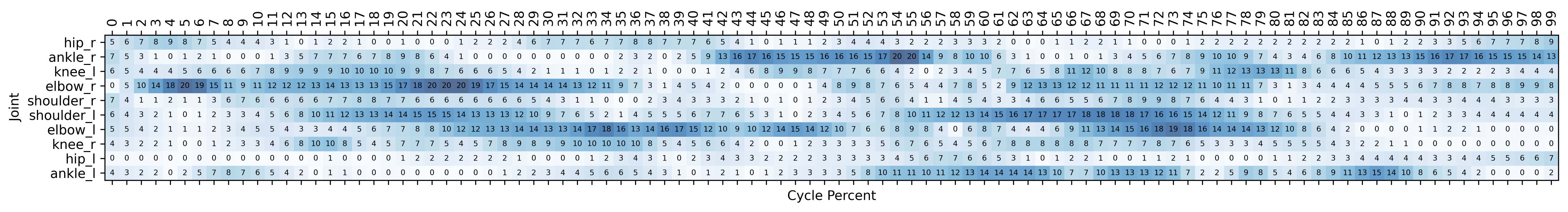} 
    \caption{Degree of deviation from typical for test video \textbf{four from CASD}~\cite{Ranjan2024GAVD} with \textbf{atypical} gait pattern. Dark regions indicating greater degree of deviation from mean angle for cycle percent}
\vspace{-10pt}
    \label{fig:vid_4_matrix}
\end{figure*}

\subsubsection{Severity of Abnormality}
The severity of abnormality, as measured by deviation from the mean, is a useful feature to observe for clinical analysis of movement~\cite{whittle2014gait}. To facilitate this, we visualise the degree of abnormality deviation from the mean. This visualisation, displayed as the degree of shading per joint over the gait cycle, represents greater issues of abnormality with darker regions. 
For example, Fig.~\ref{fig:vid_1_matrix}, which is a visualisation of test video 1 with a typical gait pattern, demonstrates fewer shaded regions. 
In contrast, Fig.~\ref{fig:vid_2_matrix}, a visualisation of test video 2 with an atypical gait pattern, shows more shaded regions. 
Similarly, Fig.~\ref{fig:vid_3_matrix} shows increased changes compared to Fig.~\ref{fig:vid_4_matrix}, which feature the same subject performing typical and atypical gait patterns, respectively. This visualisation can highlight changes in an individual's gait patterns. 

As a result, the severity of abnormality can be effectively visualised. This feature is important for prioritising healthcare focus for clinicians and serves as a tool to measure changes in deviation for healthcare management. 

\subsubsection{Qualitative Visualisation}
\begin{figure*}[!h]
    \centering
    \includegraphics[width=0.8\textwidth]{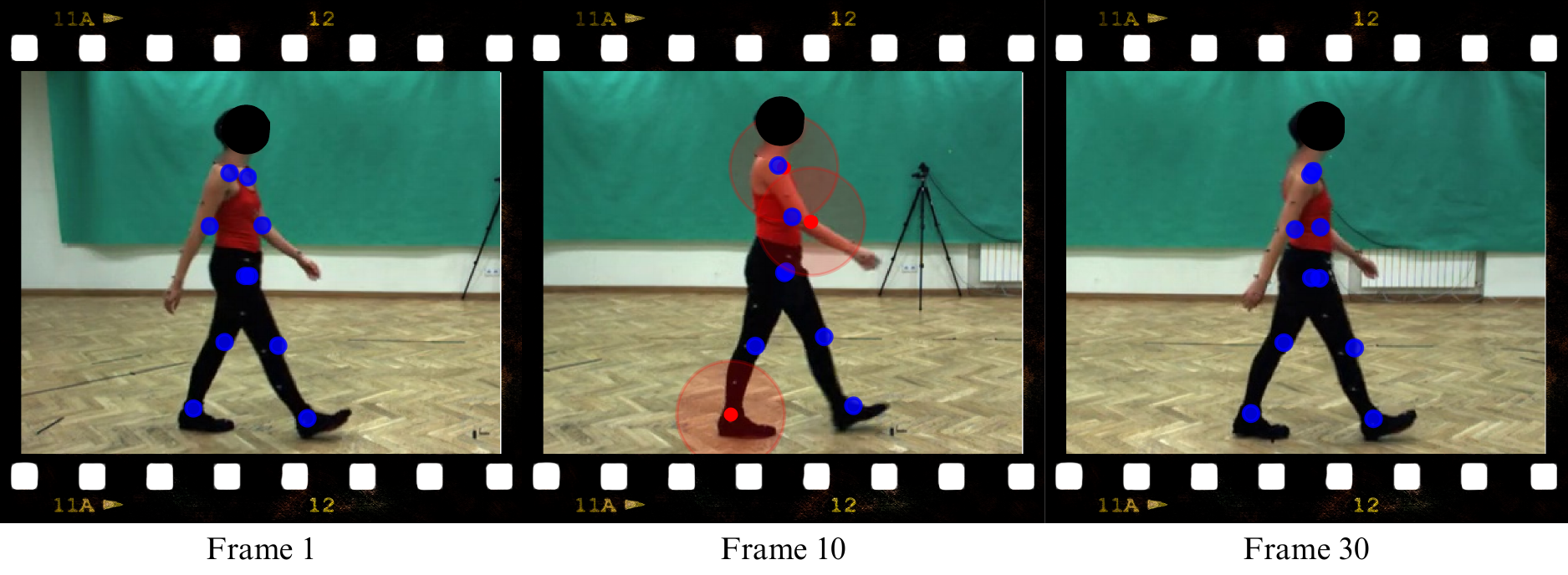} 
    \caption{Sequence of frames showing visualisation of potential gait abnormality in example with ~\textbf{normal gait pattern}. Blue indicates joint angle is within 1 standard deviation of mean. Red indicates joint angle is outside of 1 standard deviation of mean.}
    \label{fig:normal_seq}
\vspace{-6pt}
\end{figure*}

\begin{figure*}[!h]
    \centering
    \includegraphics[width=0.8\textwidth]{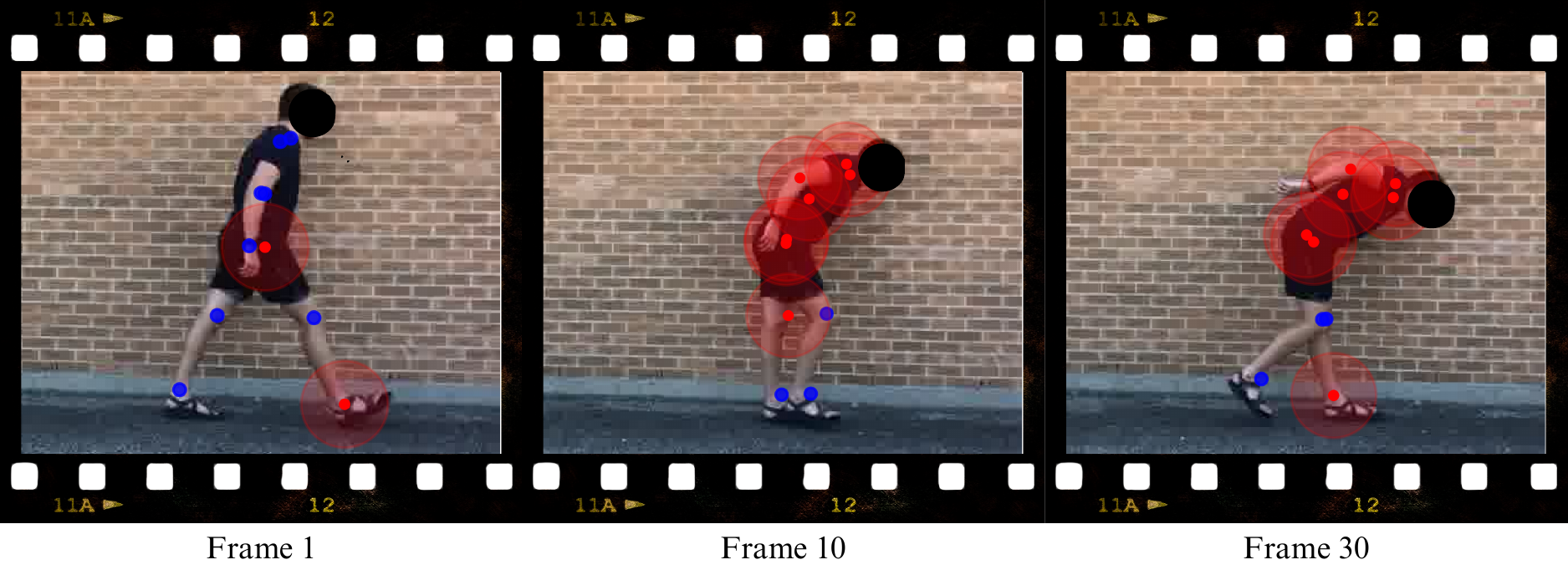} 
    \caption{Sequence of frames showing visualisation of potential gait abnormality in example with ~\textbf{abnormal gait pattern}. Blue indicates joint angle is within 1 standard deviation of mean. Red indicates joint angle is outside of standard deviation of mean.}
    \label{fig:abnormal_seq}
\vspace{-6pt}
\end{figure*}

We provide qualitative visual analysis of multiple joint abnormality detection during the gait cycle. Fig.~\ref{fig:normal_seq} and Fig.~\ref{fig:abnormal_seq} demonstrate example frames from typical and atypical gait cycle sequences, respectively, showing gait abnormalities. During the gait cycle, potential joint angle abnormalities are visualised in red over the corresponding joints. This qualitative visualisation allows clinicians to monitor the accuracy of pose estimation detection and observe deviations in gait. Such visualisations are suitable for use in patient education, medical history records, and information for treatment decision-making.

\subsection{Limitations and Future Work}\label{subsect:limits_future_work}
This study lays the groundwork for advancements in gait and action analysis using accessible methods like video recording. While progress has been made, several limitations highlight areas for future research.

One limitation is the absence of camera calibration, which could reduce distortions and parallax errors, improving the accuracy of joint position estimates. Additionally, refining joint angle selection for specific actions is crucial. For example, while upper limb joints vary significantly in gait, they could offer insights like detecting object carrying. Conversely, tasks like throwing may require prioritising upper limb joints.

The limited camera angles used in this study also restrict data comprehensiveness. Future work should incorporate multiple perspectives to enhance movement analysis accuracy. Moreover, the uncontrolled nature of the dataset, while increasing representation, challenges the balance between generalisability and accuracy.

Finally, exploring alternative pose estimation models could improve accuracy in capturing human motion. Comparative studies are needed to evaluate their impact on analysis outcomes.

\section{Conclusions}

In this paper, we have shown a feasible data-driven approach to developing normative kinematic values for gait action analysis. Our approach addresses current limitations in gait analysis using computer vision by offering cycle-wise detection of gait abnormalities, which is crucial for clinical experts such as physiotherapists. We have demonstrated the detection and highlighting of potential abnormalities in both single and multiple joints, supported by qualitative visualisations that enhance explainability. This work establishes a baseline for future advancements in gait analysis using computer vision. 


\section*{Acknowledgment}
The development of the baseline analysis technique was supported in part by an Australian Research Council (ARC) Discovery Project awarded to JK (DP230100303).

\bibliographystyle{IEEEbib}
\bibliography{main}

\vspace{12pt}
\end{document}